\documentclass{ametsocV6.1}
\usepackage[utf8]{inputenc}
\usepackage{datetime}
\usepackage{url}
\usepackage{subfig}
\usepackage{float}
\usepackage{setspace}
\usepackage{graphicx}
\usepackage{hyperref}
\usepackage{amsmath}
\usepackage{longtable}
\usepackage{caption}
\usepackage{array}
\usepackage{tabularx}
\usepackage{booktabs}

\title{ Multi-decadal Sea Level Prediction using Neural Networks and Spectral Clustering on Climate Model Large Ensembles and Satellite Altimeter Data}

\authors{Saumya Sinha,\aff{a}\correspondingauthor{Saumya Sinha, saumya.sinha@colorado.edu} 
John Fasullo,\aff{b} 
R. Steven Nerem,\aff{a} 
Claire Monteleoni\aff{a,c}
}

\affiliation{\aff{a}{ University of Colorado, Boulder, CO, USA}\\
\aff{b}{National Center for Atmospheric Research, Boulder, CO, USA}\\
\aff{c}{INRIA, Paris, France}
}

\abstract{Sea surface height observations provided by satellite altimetry since 1993 show a rising rate (3.4 mm/year) for global mean sea level. While on average, sea level has risen 10 cm over the last 30 years, there is considerable regional variation in the sea level change. Through this work, we predict sea level trends 30 years into the future at a 2-degree spatial resolution and investigate the future patterns of the sea level change. We show the potential of machine learning (ML) in this challenging application of long-term sea level forecasting over the global ocean. Our approach incorporates sea level data from both altimeter observations and climate model simulations. We develop a supervised learning framework using fully connected neural networks (FCNNs) that can predict the sea level trend based on climate model projections. Alongside this, our method provides uncertainty estimates associated with the ML prediction. We also show the effectiveness of partitioning our spatial dataset and learning a dedicated ML model for each segmented region. We compare two partitioning strategies: one achieved using domain knowledge, and the other employing spectral clustering.  Our results demonstrate that segmenting the spatial dataset with spectral clustering improves the ML predictions.} 

\begin{document}

\maketitle

%
%
%
\statement
Long-term projections are needed to help coastal communities adapt to sea level rise. Forecasting multi-decadal sea level change is a complex problem. In this paper, we show the promise of machine learning in producing such forecasts 30 years in advance and over the global ocean. Continued improvements in prediction skills that build on this work will be vital in sea level rise adaptation efforts.
%
%

%
\section{Introduction}
Satellite altimeter observations since 1993 indicate that the global mean sea level is rising at a rate of 3.4 mm/year and accelerating by 0.08mm/year$^2$, as shown in studies~\citep{nerem2018climate,hamlington2020origin}. Global mean sea level has risen 10 cm in the last 30 years. However, there is considerable regional variation in the amount of sea level rise~\citep{hamlington2016ongoing} necessitating the need for a regional sea level change analysis. With three decades of satellite observations, we can now investigate the role played by anthropogenic climate change signals such as greenhouse gasses, aerosols, and biomass burning in this rising sea level. Climate model projections can be used to estimate the extent of the causal contributions from such factors and forecast future sea level changes. In~\citep{fasullo2018altimeter,fasullo2020sea,fasullo2020forced}, two large ensembles of climate models were studied to show that the forced responses to greenhouse gas and aerosols have begun to emerge in the regional pattern of sea level rise in the altimeter data. This motivates us to utilize climate models in our framework. Our work uses machine learning to predict future regional patterns of sea level change. It is part of a longer-term research project that investigates the extent of contributions from anthropogenic climate-change signals to sea level change. Through our work, we show promising results demonstrating the potential of neural network-based ML models. Our framework uses both satellite observations and climate model simulations to predict sea level trends 30 years into the future at a 2-degree spatial resolution.

Forecasting long-term sea level change is a complex problem given the natural variability of the ocean, the wide range of processes involved, and the complex non-linear interactions playing a role in sea level change. Some past studies have used satellite altimeter data and adopted ML techniques to perform sea level prediction. Tide-gauge data has also been used for similar tasks but it suffers from the influence of local coastal effects and poor spatial coverage, while satellite altimeter data provides nearly global coverage. ~\citet{imani2017spatiotemporal} make use of support vector regression for sea level prediction in the tropical Pacific Ocean. In~\citep{braakmann2017sea}, they utilize a combination of CNN (Convolutional neural network) + ConvLSTM~\citep{shi2015convolutional} layers to perform interannual sea level anomalies (SLA) prediction over the Pacific Ocean.~\citet{zhao2019sea} use a combination of least squares and neural networks to produce sea level anomalies prediction in the Yellow Sea.~\citet{sun2020estimation} work with LSTM (Long short-term memory network) for the South China Sea. Through their work in~\citep{liu2020sea}, the authors employ an attention-based LSTM mechanism for sea surface height (SSH) forecasting in the South China Sea.~\citet{balogun2021sea} include ocean-atmospheric features like sea surface temperature, salinity, and surface atmospheric pressure to build support vector and LSTM models for the West Peninsular Malaysia coastline.~\citet{nieves2021predicting} make use of gaussian processes and LSTM to predict sea level variation along the regional coastal zones. In~\citep{hassan2021comparative}, they compare various machine learning techniques to predict global mean sea level rise. An important part of the pipeline in~\citep{AHybridMultivariateDeepLearningNetworkforMultistepAheadSeaLevelAnomalyForecasting} includes a ConvLSTM pipeline consisting of 3D convolutions and attention modules for forecasting altimeter SLA on the South China Sea. These techniques, however, are trained only on the altimeter dataset which to date is only 30 years in length, this can affect the performance of such data-driven models as brought up in this latest survey by~\citet{bahari2023predicting}. These approaches also do not use the insights provided by climate model projections that can potentially inform on contributions of anthropogenic climate-change signals. Moreover, these models address regional forecasting with a lead time of a few days to a few years ahead, but do not go so far as to forecast sea level change over the global ocean 30 years in advance. Our work utilizes the climate model projections and addresses the problem at a much bigger spatial scale that includes all the oceans and a much longer time horizon in the future.

We work with 30-year linear trends of the sea level time series. We note that the climate models do not accurately reproduce all aspects of the trend pattern in altimeter data. There is also more variability in the altimeter trends compared to the climate models. We observed this in our previous work~\citep{2022AGUFMOS36A..05S}, where a UNet~\citep{ronneberger2015u} model is trained on long periods of climate model simulations to produce spatiotemporal predictions 30 years ahead. This UNet model is then used to predict the future altimeter data. However, these predictions had much lower variability as compared to the altimeter observations. This underscores the challenge of combining modeled and observed fields in producing sea level predictions.

Working with multi-decadal global trends severely limits the ground-truth data we have. Thus we use the sea level trend values at every spatial grid point to create a training dataset for our ML model. With a 2-degree spatial resolution, we get a 180x90 (longitude x latitude) grid in our sea level trend maps. This gives us a reasonably large dataset for training an ML model even for a single 30-year-long trend for each grid point. We build a supervised learning framework using fully connected neural networks (FCNNs) that learns a non-linear mapping of the climate model trends to predict the altimeter trend while absorbing the biases that the climate models have away from the altimeter observations. This is accompanied by an interpretability study that explains the contributions of all the climate models to our final prediction. Given that the dominant factors driving sea level variability differ by region, we segment our spatial dataset and learn separate FCNNs for each segmented region. We compare a partition achieved using domain knowledge to a partition achieved via spectral clustering. We show that segmenting the spatial dataset improves the ML predictions. Spectral clustering shows promise by predicting future trends with ML such that their variability lies in the range we expect, given the variability of the past altimeter observations. Our predictions with spectral clustering also have lower uncertainties in impactful areas.

 \section{Method}
\label{method}
\begin{figure}[h]
\begin{center}
\includegraphics[scale=0.3]{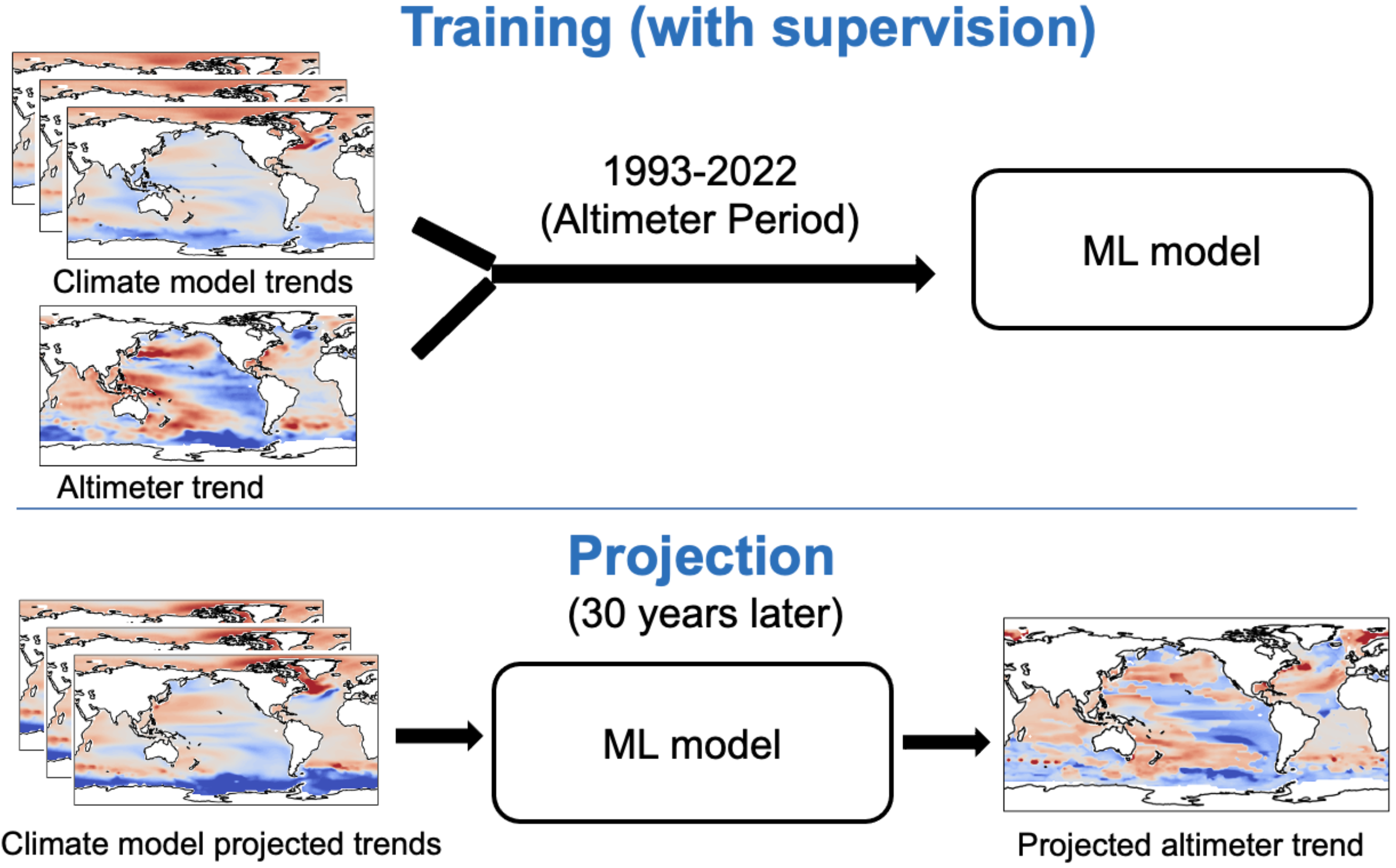}
\end{center}
\caption{Overall machine learning pipeline for the task of sea level trend prediction using trends from climate models projections and altimeter observations.}
\label{label:trend_ML_pipeline}
\end{figure}

Our supervised learning pipeline is trained for the period 1993-2022. The spatial grid is flattened to create our dataset, where each data point corresponds to an oceanic grid point.  For every grid point, a linear trend is computed over 1993-2022 for the climate model ensemble means, (described in~\ref{dataset}), comprising the input features (X). The trends computed for the altimeter data (ground truth) make the label (Y) for our supervised ML training. So, we get the supervision from the altimeter trend, and the features to our ML model from the climate model hindcast trends. In the inference phase, we predict trends for 30 years later. This is done by taking the climate model projected trends for 2023-2052 and passing them through the learned ML model to predict the altimeter trend. See the overall ML framework in Figure~\ref{label:trend_ML_pipeline}. The ML model is a fully connected neural network (FCNN) trained with mean squared error (MSE) as the loss. The MSE is weighted, where the weights are the cosine of the latitude of the grid points. This gives spatial-weighting which essentially assigns more weight to the equatorial regions and less weight to polar regions. 

\begin{figure}[!htb]
 \centering
 \subfloat[\large Spectral clustering]{%
      \includegraphics[width=0.45\textwidth]{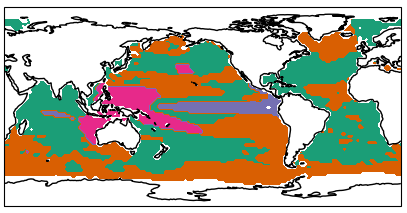}}
 \qquad
 \subfloat[\large Domain-specified partition]{%
      \includegraphics[width=0.45\textwidth]{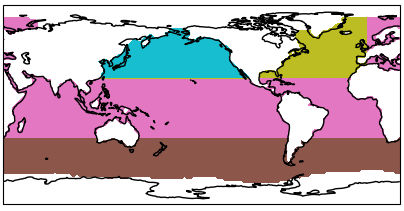}}
\caption{Spatial segmentations obtained from (a) Spectral clustering, and (b) a Domain-specified partition derived from our physical understanding of the data, where the North Atlantic Ocean (olive) and North Pacific Ocean (cyan) are assigned individual partitions, latitudes from south up to -30 is assigned another partition (brown), and the rest belong to the 4th partition (pink).}
\label{label:clusters}
\end{figure}

\textbf{Clustering}:
We segment our spatial grid into partitions or clusters and observe the performance of the ML model when trained based on these clusters i.e. a separate FCNN is trained for \emph{each} cluster. This is based on the hypothesis that learning ML model weights that are attuned to each cluster can be more optimal than a single ML model for the entire globe. Our study compares Spectral clustering against a Domain-specified partitioning that is derived from our physical knowledge of the data and proposed by the domain experts in the team. The time series of the altimeter sea-surface height (with the seasonality removed) serves as the features for Spectral clustering. Empirical evaluation with K-means clustering failed to perform close to Spectral clustering and is not included in the study. The spatial segmentations with Spectral clustering as well as the Domain-specified partition can be seen in Figure~\ref{label:clusters}. The Spectral clustering, as observed by domain experts, seems to be influenced by the ENSO (El Niño–Southern Oscillation) phenomenon in the Pacific region. This could be because of the similarity between spectral decomposition and EOF (Empirical orthogonal function) analysis and the fact that ENSO is the leading mode of interannual climate variability~\citep{vestergaard2010seventeen}. This could be beneficial as creating these clusters helps to treat ENSO-specific regions separately. These partitioning strategies are compared to each other and to a setup where the spatial grid is not segmented at all. 

We make use of k-fold cross-validation (k=5) to choose the best hyperparameters for each cluster, ending up with different FCNN architectures per cluster. To elaborate further, each of the orange and green clusters in the Spectral clustering setup as seen in Figure~\ref{label:clusters}(a), learns an FCNN consisting of 3 hidden layers with 1024, 512, and 256 neurons respectively. For each of the other two smaller clusters, we use an FCNN with 2 hidden layers and 256, and 128 neurons respectively. Each hidden layer is followed by a \textit{relu} activation. \textit{l2}($0.000005$) regularizer and a single dropout layer ($0.2$) are applied to avoid overfitting in each of the ML models. 

\section{Dataset}
\label{dataset}
\begin{figure}[h]
\begin{center}
\includegraphics[scale=0.25]{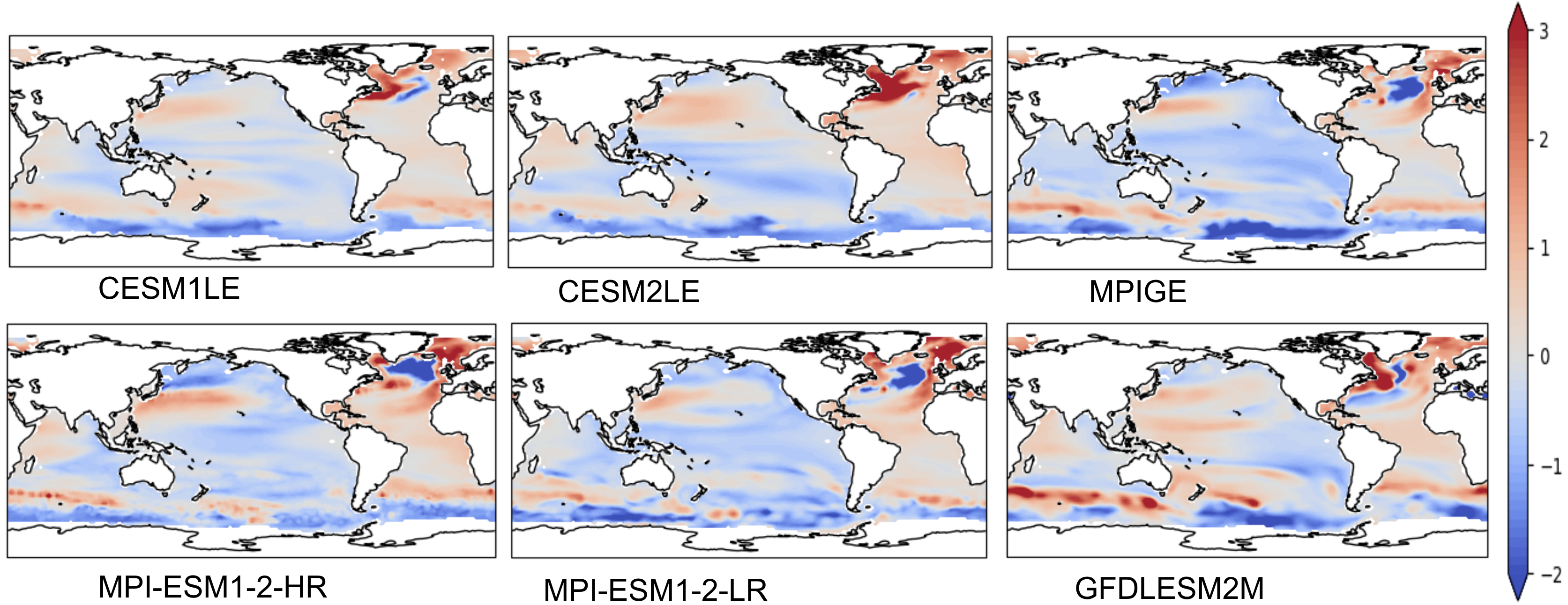}
\end{center}
\caption{Figure shows the sea level trend maps for the six climate model large ensembles for the period 1993-2022 with their global mean removed. Here the trend values are visualized in mm/year.}
\label{label:trend_climate_models}
\end{figure}
\begin{figure}[h]
\begin{center}
\includegraphics[scale=0.85]{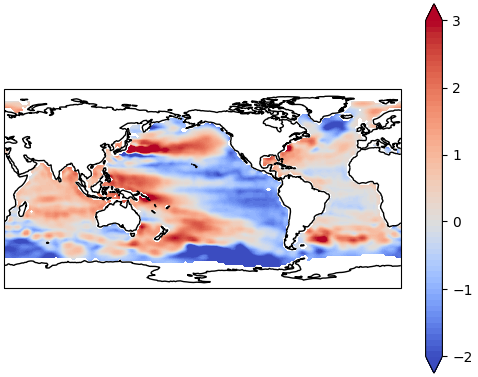}
\end{center}
\caption{Figure shows the altimeter sea level trend map for the period 1993-2022 with the global mean removed. Here the trend values are visualized in mm/year.}
\label{label:trend_altimeter}
\end{figure}
Two types of data are used in this study: altimeter data and climate model large ensemble (LE) experiments. The altimeter dataset is a monthly sea surface height (SSH) data at 1/4-degree spatial resolution for the time period 1993-2022. A spatialsmoothing\footnote{\url{https://www.ncl.ucar.edu/Document/Functions/Built-in/exp_tapersh.shtml}} is applied to reduce the influence of small-scale ocean eddies. For the same duration, we obtain monthly SSH at 1-degree spatial resolution from the ensemble means of six different climate model LEs produced with CESM1~\citep{kay2015community}, CESM2~\citep{danabasoglu2020community}, GFDLESM2M~\citep{dunne2013gfdl}, MPIGE~\citep{maher2019max}, MPI-ESM1-2-HR~\citep{muller2018higher} and MPI-ESM1-2-LR~\citep{giorgetta2013climate}. These LEs provide simulations for the 20th and 21st centuries and are multi-member ensembles of climate models run with small perturbations in the initial conditions to estimate the distribution of internal climate variability and forced climate change. Model simulations for individual members of the above large ensembles are averaged to create the sea surface height (SSH) variable. We do this since we expect internal variability to be inherently unpredictable while we expect the response to external forcings (influences considered external to the climate system that impact climate) to be both predictable and slowly varying. This step removes noise but it also reduces variability in the climate model SSH data, while we have variability present in the altimeter SSH data as it is a single field. 

Some of the climate models operate under assumptions in which their global mean is by definition 0. We, therefore, remove the global mean in all the datasets including the altimeter. The spatial SSH fields for both the altimeter data and climate model output are regridded to a 2 degree, i.e. a 180x90 grid as it speeds up the computation while still keeping a reasonable resolution.  For every ocean grid point, a linear trend is fitted to the monthly SSH time series for the 1993-2022 (30-year) time period. This way, a \textit{single} trend map is obtained, for all the ensembles and the altimeter (see Figures~\ref{label:trend_climate_models} and~\ref{label:trend_altimeter}). Working with trends helps to avoid the monthly variability of the SSH fields. We can observe the differences between the altimeter and the climate model trends, especially with respect to variability. The climate models do not accurately reproduce the trend pattern in altimeter data and there is a lot more variability in the altimeter trend as compared to the climate model trends.

It is worth noting that altimeter records are not present for all latitudes. We have both altimeter and climate model trend values for $8,001$ global ocean points (excluding land grid points) that we use as the dataset for ML. We show trend values from the six climate models in Figure~\ref{label:trend_climate_models} that serve as the input features for our ML model and the altimeter trend value in Figure~\ref{label:trend_altimeter} that serves as the label for training the ML model. These trend values are computed in cm/year and are normalized by scaling them between 0 and 1 for training. After training, in the inference phase, trends are predicted for 30 years later. Climate model projected trends are computed in the same way for 30 years later, i.e. for 2023-2052. These are then passed through the learned ML model to predict the altimeter trend for 2023-2052.


\section{Results}
\label{results}
We report our results using different evaluation metrics for the past and future time periods, since there is no ground truth with which to evaluate future predictions.
\subsection{1993-2022}
With the ground truth data available for this period, in Table~\ref{results-table}, we report the RMSE scores on the historical (training) time period, for the two spatial segmentation strategies, compared to applying our supervised learning step directly to the entire spatial extent (\textit{No clustering}). Table~\ref{results-table} also shows the Pearson correlation scores between the ML predicted trend and the true altimeter trend for 1993-2022. 
The RMSE and correlation scores are spatially-weighted as described in Section~\ref{method}. The \textit{Domain-specified partition} is observed to have better scores (lower RMSE and higher correlation) for the training period as compared to \textit{Spectral clustering}.
Both the \textit{Domain-specified partition} and \textit{Spectral clustering} scores are considerably better than the \textit{No clustering} setup. 
Each of the segmented regions is examined by looking at each cluster's RMSE and correlation scores. The trend predicted by ML is visualized and higher error zones are mostly observed in the green cluster of Figure~\ref{label:clusters}(a) for \textit{Spectral clustering} and the olive cluster of Figure~\ref{label:clusters}(b) for \textit{Domain-specified partition}. 
\begin{table}[t]
\begin{center}
\begin{tabular}{lll}
\multicolumn{1}{c}{\bf Method}  &\multicolumn{1}{c}{\bf RMSE$\downarrow$} &\multicolumn{1}{c}{\bf Correlation$\uparrow$}
\\ \hline \\
No Clustering          & 0.72 & 0.82 \\
Domain-specified partition           & 0.4 & 0.95 \\
Spectral Clustering             & 0.51 & 0.91 \\
\end{tabular}
\end{center}
\caption{Table comparing the ML prediction performance in terms of weighted RMSE (in mm/year) and correlation for different spatial segmentations for 1993-2022.}
\label{results-table}
\end{table}
While these scores explain the ML's training performance, our interest mainly lies in the future period prediction which is detailed below.

\subsection{2023-2052}
It is harder to gauge the performance of any ML method without the ground truth. In this case, we do a qualitative analysis of the predicted trend in terms of cumulative variability, to evaluate the ability of the ML models to predict trends with variability similar to the variability of the 1993-2022 altimeter trend. Additionally, we compute the model uncertainty of the ML models in their prediction. As often done in the climate science domain, we also evaluate the ML models solely with the climate model datasets~\citep{monteleoni2011tracking}. These experiments are described later in this section. 
\begin{figure}[htbp]
  \centering
\begin{subfloat}[\large ML predicted trend with no partitioning for 2023-2052 (RMS: 0.81mm/year).]
{\includegraphics[width=0.65\textwidth]{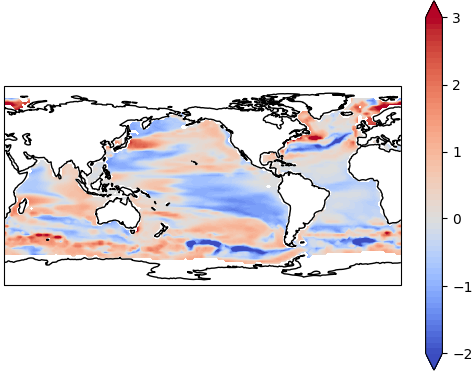}}
  \end{subfloat}
  \begin{subfloat}[\large ML predicted trend with the Domain-specified partition for 2023-2052 (RMS: 1.68mm/year).]
{\includegraphics[width=0.65\textwidth]{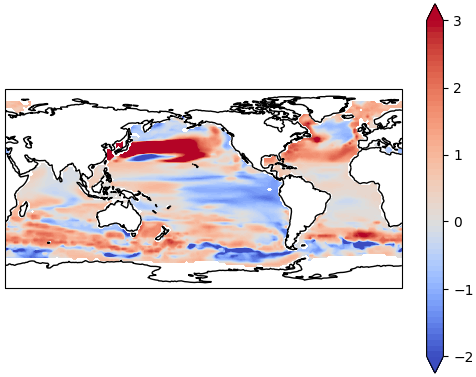}}
  \end{subfloat}
  \caption{}
  \end{figure}
  \clearpage
  \begin{figure}[!t]\ContinuedFloat
    \centering
 \begin{subfloat}[\large ML predicted trend with Spectral clustering for 2023-2052 (RMS: 1.05 mm/year).]
{\includegraphics[width=0.65\textwidth]{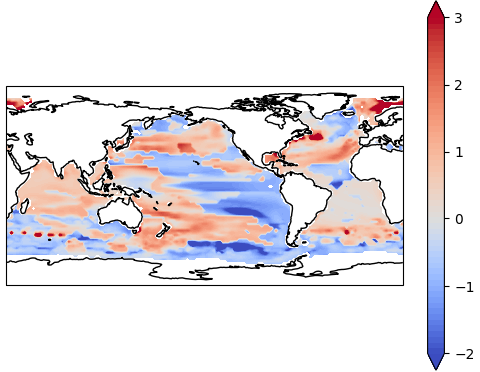}}
  \end{subfloat}
  \caption{The trend estimates are predictions for the future period: 2023-2052 in mm/year. Figures show the trend predicted with ML using (a) no partitioning of the spatial grid, (b) the Domain-specified partition, and (c) Spectral clustering.}
  \label{label:prediction_outcomes}
\end{figure}

We use the root mean square (RMS) of the trend (spatially-weighted as in Section~\ref{method}) to quantify the notion of variability in the trend. The RMS value is higher if the cumulative variability is higher and vice-versa. Figure~\ref{label:trend_altimeter} shows that the altimeter trend from 1993-2022 
has a high variability. We computed the RMS value to be \textbf{1.23mm/year}. This gives us the baseline variability of \textit{persistence}, a standard baseline approach in climate and weather forecasting, i.e., considering this observed variability as an estimate of future variability.
In Figures~\ref{label:prediction_outcomes}(a), (b), and (c), we show the future predicted trend obtained from the ML model without any partitioning (\textit{No clustering}), with the \textit{Domain-specified partition}, and with a partition obtained via~\textit{Spectral clustering}, respectively. We computed the RMS values associated with trend predictions obtained from the three strategies. Trend predicted with \textit{Spectral clustering} (Figure~\ref{label:prediction_outcomes}(c)) shows a high variability with RMS as \textbf{1.05 mm/year} for 2023-2052. It is very close, though still slightly less than the altimeter trend variability of the past. On the other hand, the trend predicted with the \textit{Domain-specified partition} (Figure~\ref{label:prediction_outcomes}(b)) shows a much higher variability with RMS as \textbf{1.68mm/year}. The high prediction red zone in the North Pacific Ocean could be dominating the overall RMS value of the \textit{Domain-specified} prediction. This emphasizes the need to use additional metrics and analyses to evaluate our predictions rather than relying on a single overall score expressing cumulative variability.  
Notably, the predicted variability of both \textit{Spectral clustering} and the \textit{Domain-specified partition} are higher as compared to the \textit{No clustering} setting (RMS: \textbf{0.81mm/year}). This result strengthens our hypothesis that segmenting the spatial grid and learning one ML model 
on each segmented region yields predictions that can better capture variability (with respect to persistence). Measuring the 
correlation between the persistence and future predicted trend is also useful as it is expected to be fairly high based on the climate model experiments which also show a high correlation between the past and future trend in their projections. This correlation is much higher (\textbf{0.59}) for \textit{Spectral clustering} than the \textit{Domain-specified partition} (\textbf{0.45}) and \textit{No clustering} setup (\textbf{0.27}).

\subsubsection{Model uncertainty} 
Providing uncertainties of machine learning predictions can be extremely useful. For this application, we do so as another way to evaluate our ML model's future predictions.
~\citet{gal2016dropout} 
showed theoretically that neural networks with dropout layers can be interpreted as a Bayesian approximation of a deep Gaussian process. 
Thus we can obtain uncertainties with dropout neural networks without sacrificing accuracy and with lesser computation cost as compared to the Bayesian models. 
\begin{figure}[htbp]
  \centering
  \begin{subfloat}[\large ML model uncertainty in terms of standard deviation (mm/year) over future prediction with the Domain-specified partition.]
{\includegraphics[width=0.75\textwidth]{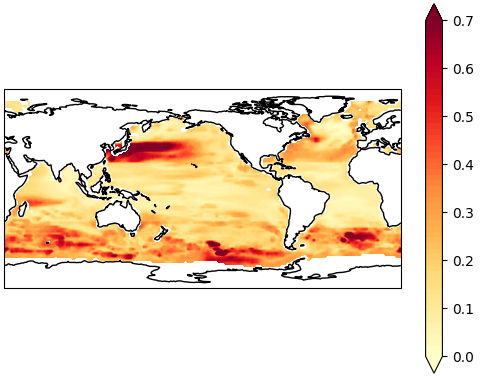}}
  \end{subfloat}
  \begin{subfloat}[\large ML model uncertainty in terms of the standard deviation (mm/year) over future prediction with Spectral clustering.]
{\includegraphics[width=0.75\textwidth]{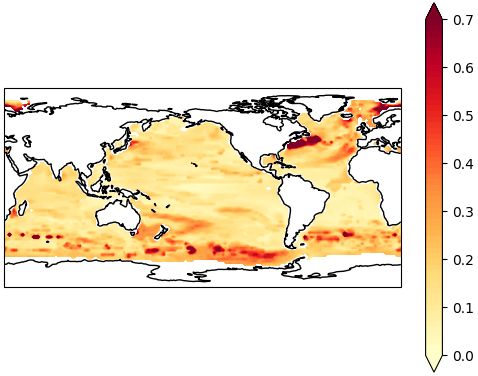}}
  \end{subfloat}
  \caption{Figure shows ML model uncertainty map over future prediction with (a) the Domain-specified partition, and (b) Spectral clustering.}
  \label{label:uncertainty}
\end{figure}
This Monte Carlo dropout approach can work with any existing neural networks trained with dropout~\citep{gal2016dropout}. 

Our FCNN model includes dropout layers to reduce overfitting while training, thus allowing us to use the Monte Carlo dropout approach for uncertainty estimation. To do so, in the inference phase, we perform multiple forward passes (with different dropout masks) through our ML model. We then report the mean of the ensemble of predictions as the prediction outcome and their standard deviation as the prediction uncertainty. 
Figure~\ref{label:uncertainty} shows the prediction uncertainty plots for both \textit{Spectral clustering} and the \textit{Domain-specified partition}. The predictions with the \textit{Domain-specified partition} (Figure~\ref{label:uncertainty}(a)) show a higher overall variance in prediction. We observe higher uncertainties in key areas that are critical for socio-economic impacts such as important parts of the Pacific Ocean whereas \textit{Spectral clustering} (Figure~\ref{label:uncertainty}(b)) predictions are more confident in most of the Pacific Ocean and higher uncertainties are concentrated in the Southern Ocean and parts of the North Atlantic Ocean. 
We also studied the cumulative uncertainty by taking the root mean square (RMS) of this model uncertainty over the global ocean. Lower RMS is better as it indicates lower cumulative uncertainty. The RMS for \textit{Spectral clustering}  (\textbf{0.19mm/year}) is better than the \textit{Domain-specified partition} (\textbf{0.24mm/year}) and the \textit{No clustering} scenario (\textbf{0.3mm/year}). We observe that the ML model is more certain with \textit{Spectral clustering}.

We also observed that some regions over which the ML model with \textit{Spectral clustering} had higher uncertainties (Figure~\ref{label:uncertainty}(b)) had high overlap with the regions where climate model projections for 2023-2052 had the highest disagreement (Figure~\ref{label:future_climate_models}(b)).


\subsubsection{Interpretability Study} 
\label{Interpretability}
 Through this interpretability study, our goal was to understand the contribution of each climate model in the ML prediction. While complex machine learning models can predict accurate outcomes, it is extremely important to understand why the ML model makes a certain prediction in order to make it more interpretable. We use SHAP (SHapley Additive exPlanation)~\citet{lundberg2017unified} to compute the contributions of each feature to a prediction outcome in order to explain the prediction.
 
 \citet{lundberg2017unified} in their work on SHAP show the value of a linear explanation model that is an interpretable approximation to the original complex model by proposing a class of methods: \textit{Additive feature attribution methods}. They use $x$ and $f$ as the original inputs and prediction model, $g$ as an explanation model, and $x^\prime$ as a simplified input such that $x = h_x(x^\prime)$, $h_x$ being a mapping function. Under the definition of \textit{Additive feature attribution methods} as described in~\citep{lundberg2017unified}, the explanation model $g$ must satisfy $g(z^\prime) \approx f(h_x(z^\prime))$ where $z^\prime \approx x^\prime$, and can be written as $g(z^\prime)=\phi_0 + \sum_{n=1}^{M}\phi_iz_i^\prime$, $z^\prime \in \{0,1\}^M$ indicates whether a particular feature (out of $M$ features) is included or not with a binary value. Here, $\phi_i$ indicates feature attribution or feature importance i.e. how much this feature contributed to the model's outcome. Lundberg et al. then show from the game theory literature that Shapley values as $\phi_i$ satisfy the definition and a few more desirable properties of this class of methods~\citep{shapley1953value,Young1985MonotonicSO}. For the computation of Shapley values~\citep{lipovetsky2001analysis}, marginal contribution of a feature $i$ is computed by taking the difference between the model $f$'s output with and without that feature. The marginal contribution is computed for all possible subsets $S \subseteq F \setminus \{i\}$ ($F$ is the set of all features), and a weighted average over them gives the Shapley value as shown below (from \cite{lundberg2017unified}).
\begin{equation}
 \phi_i =  \sum_{S \subseteq F\setminus \{i\}} \frac{|S|!(|F|-|S|-1)!}{|F|!}[f_{S\cup \{i\}} (x_{S\cup \{i\}}) - f_S (x_S)]
 \end{equation}
In most cases, ML models cannot handle missing features, so this is often approximated by integrating out the feature using samples from a background dataset as discussed in\mbox{~\citep{vstrumbelj2014explaining,lundberg2017unified}}.
The computation of SHAP i.e. SHapley Additive exPlanation becomes very challenging as number of features increase. In~\citep{lundberg2017unified} they provide an approximation to obtaining the Shapley values via Kernel SHAP (a model agnostic approximation). 
\begin{figure}[htbp]
\hspace*{5cm} \begin{subfloat}[]
 {\includegraphics[width=0.4\textwidth, scale=0.25]{spectral_clustering.png}}
  \end{subfloat}
  \\
   \begin{subfloat}[]
   {
   \centering
  \includegraphics[width=0.9\textwidth]{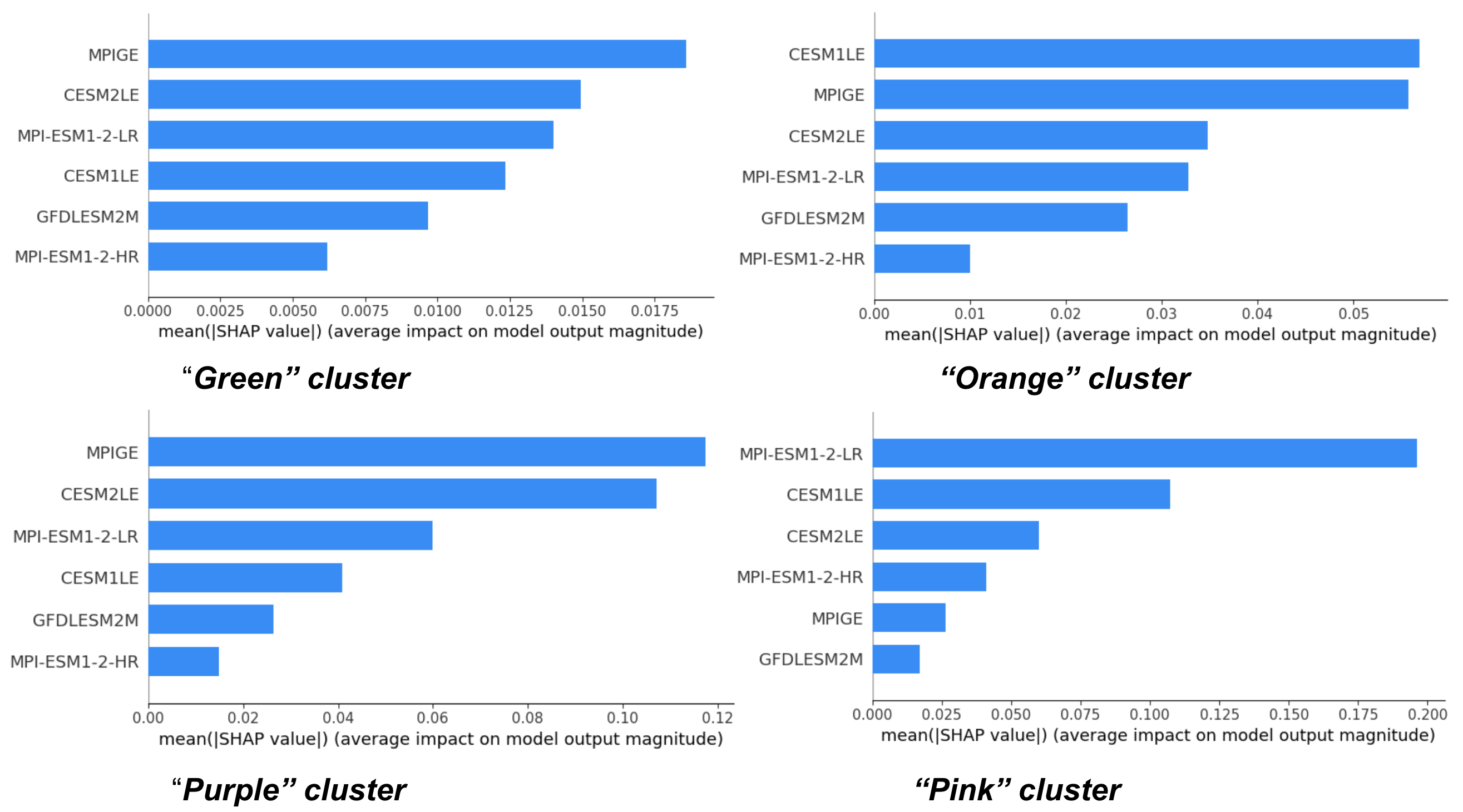} }
  \end{subfloat}
  \caption{Figure shows (a) the clusters obtained from Spectral clustering, and (b) the plot showing the feature importance ranking of all the climate models based on their contribution to the future prediction as given by SHAP values. The map with the spectral clusters is added to provide more context.}
  \label{label:shap_spectral}
\end{figure}

Our interpretability analysis is based on SHAP as explained above. We use Kernel SHAP from Python's \textit{shap.KernelExplainer} to compute the contributions or feature importance values of the climate models which are feature inputs to our FCNNs in order to explain the future prediction.
We apply SHAP on each of the clusters since we learn different ML models for different clusters. Figure~\ref{label:shap_spectral} shows a cluster level feature importance ranking of all the climate models for the \textit{Spectral clustering} setup.
SHAP assigns a feature importance to each climate model for the future prediction on 
each grid point in the cluster. These importance values are averaged over every cluster and shown in the bar plots in Figure~\ref{label:shap_spectral}. 
Given that we use scaling on the input features as well as the output labels when training the ML models, the importance values displayed in Figure~\ref{label:shap_spectral} are scaled as well.
Overall, the SHAP values indicate that the CESM1, CESM2, and MPIGE large ensembles are more important for all the clusters than others, suggesting that ML model relied more on these for future predictions. 

\subsubsection{Evaluation with climate models}
\label{leave_one_out}
\begin{figure}[htbp]
\centering
\begin{subfloat}[]
 {\includegraphics[width=1\textwidth]{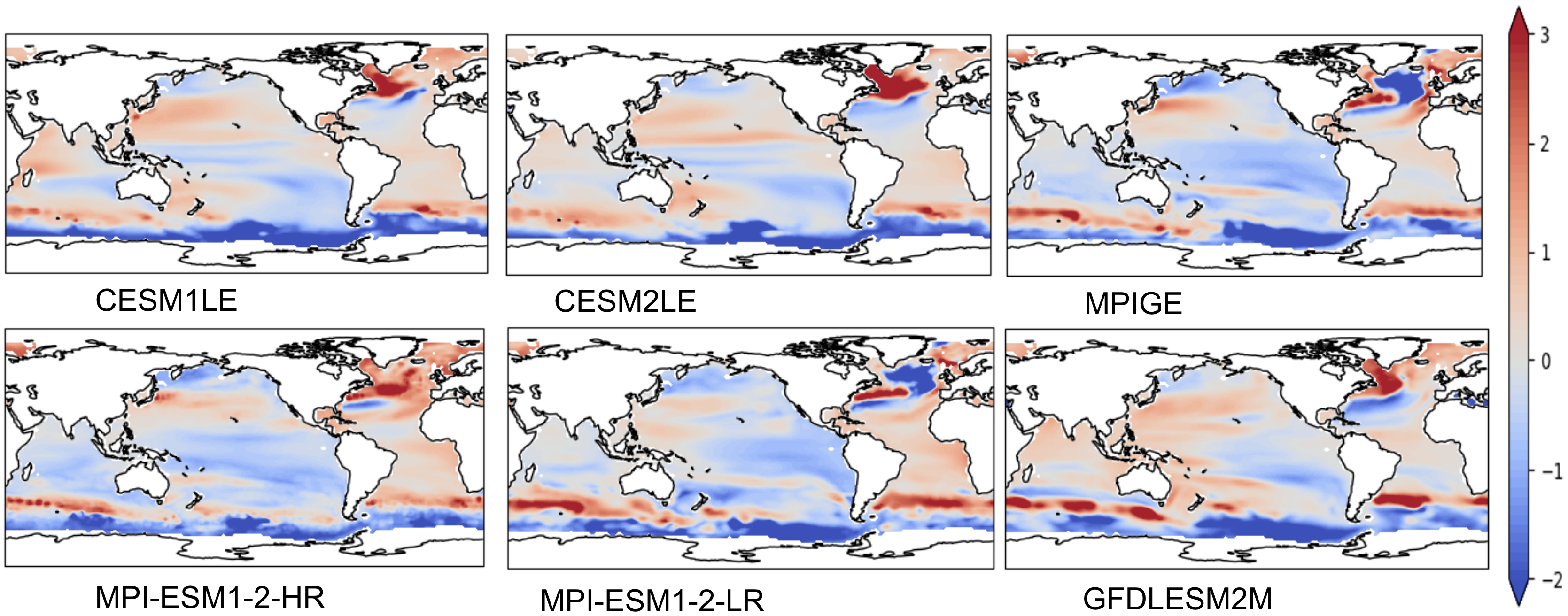}}
  \end{subfloat}
  \\
   \begin{subfloat}[]
  {\includegraphics[width=0.5\textwidth, scale=0.2]{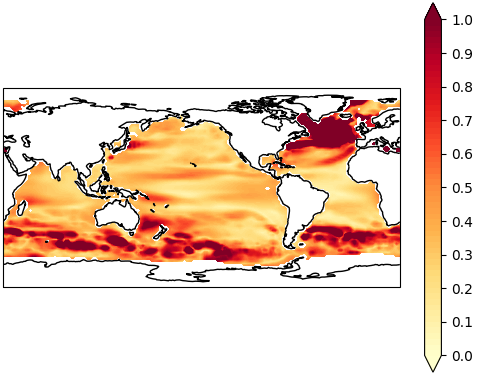} }
  \end{subfloat}
  \caption{Plot showing (a) the climate model projected trends in mm/year for 2023-2052, and (b) the standard deviation in their projections in mm/year. }
  \label{label:future_climate_models}
\end{figure}

While we do not have observations for the future to validate our prediction outcomes, we do have climate model projections for the future. As often done in the climate science domain, we perform an evaluation of our predictions using only the climate model datasets. We train the same FCNN models for 1993-2022 again, but this time using one of the climate model hindcasts as the training label instead of the altimeter data, and the rest of the five climate models as input features (like in~\citep{monteleoni2011tracking}). We train six such ML models treating each of the six climate models as the training label one at a time. At the time of inference for 2023-2052, we have the ground truth i.e. climate model projections available for the future for each climate model, so we measure RMSE and correlation scores for the ML prediction of the climate model trend against the true climate model projected trend. 
We show the weighted correlation metrics in Table~\ref{results-table_climate_models}(a) and weighted RMSE scores in Table~\ref{results-table_climate_models}(b) for both the \textit{Spectral clustering} and \textit{Domain-specified partition} setups. They are evaluated against the persistence scores for each of the climate models (here, persistence is using the climate model hindcast from 1993-2022 as the prediction for the future 2023-2052). It should be noted that the climate model projections substantially differ from each other which makes this prediction task harder for ML. Figure~\ref{label:future_climate_models} shows the trend maps from all the climate models for the future period 2023-2052 and a standard deviation plot showing the variance in their projections. 
\begin{table}[!htbp]
	\centering
\begin{subfloat}[\large Correlation$\uparrow$]
{\renewcommand{\arraystretch}{1.25}
	\begin{tabularx}{0.9\textwidth}{|X|X|X|X|X|X|X|X|}
		\cline{1-8}
		\multicolumn{1}{|c|}{} & \textbf{\small CESM1 LE} & \textbf{\small CESM2 LE} & 
		\textbf{\small MPIGE} & 
         \textbf{\small MPI-ESM1-2-HR} & 
		\textbf{\small MPI-ESM1-2-LR} &
        \textbf{\small GFDL ESM2M} &
        \textbf{\small Average}
        \\ \cline{1-8}
			{\small Persistence} & 0.74 & 0.74 & 0.73 & \textbf{0.53} & 0.49 & \textbf{0.74} & 0.66 \\ \hline
			{\small ML with Domain-specified partition} & 0.79 & 0.73 & 0.74 & 0.37 & 0.49 & 0.48 & 0.6 \\ \hline
			{\small ML with Spectral clustering} & \textbf{0.82} & \textbf{0.81} & \textbf{0.82} & 0.39 & \textbf{0.6} & 0.43 & 0.65 \\ \hline
   \end{tabularx}
  }
  \end{subfloat}
  \begin{subfloat}[\large RMSE$\downarrow$]
  {\renewcommand{\arraystretch}{1.25}
	\begin{tabularx}{0.9\textwidth}{|X|X|X|X|X|X|X|X|}
		\cline{1-8}
		\multicolumn{1}{|c|}{} & \textbf{\small CESM1 LE} & \textbf{\small CESM2 LE} & 
		\textbf{\small MPIGE} & 
         \textbf{\small MPI-ESM1-2-HR} & 
		\textbf{\small MPI-ESM1-2-LR} &
        \textbf{\small GFDL ESM2M} &
        \textbf{\small Average}
        \\ \cline{1-8}
			{\small Persistence} & 0.58 & 0.6 & 0.69 & \textbf{0.83} & 0.95 & \textbf{0.72} & 0.73 \\ \hline
			{\small ML with Domain-specified partition} & 0.56 & 0.69 & 0.73 & 0.96 & 0.99 & 0.99 & 0.82 \\ \hline
			{\small ML with Spectral clustering} & \textbf{0.49} & \textbf{0.56} & \textbf{0.58} & 0.93 & \textbf{0.86} & 1.01 & 0.74 \\ \hline
  \end{tabularx}
  }
  \end{subfloat}
\caption{An evaluation setup using climate models to simulate observation data so as to evaluate ML predictions for the future period 2023-2052. The table reports the performance score in terms of (a) correlation (higher the better) and (b) RMSE (in mm/year, lower the better) between the ML prediction of a climate model and the ground truth climate model projection. Results are computed for predicting each of the six climate models one at a time (shown as the columns in the table) using the rest of them as input features. The last column shows an average score obtained by averaging the scores over all climate models.}
\label{results-table_climate_models}
\end{table}

Based on both the correlation and RMSE scores, it can be seen that ML with \textit{Spectral clustering} outperforms the \textit{Domain-specified partition} on nearly all the climate models, falling slightly behind only for the case of GFDLESM2M. It also performs better than the persistence on all the climate models except MPI-ESM1-2-HR and GFDLESM2M. 
We observed that the regions where MPI-ESM1-2-HR and GFDLESM2M predictions with the \textit{Spectral clustering} setup have higher errors (in parts of the Southern Ocean and the North Atlantic Ocean) are some of the regions where these two climate models have disagreement over with the remaining climate model projections (see Figure~\ref{label:future_climate_models}). 
Comparing the average correlation and RMSE scores (last column in Table~\ref{results-table_climate_models}) over all the six ML models based on the six climate model labels show \textit{Spectral clustering} to be better than the \textit{Domain-specified partition} and very close to the persistence.  


\subsubsection{Experiment with varying number of clusters}
\label{varying_clusters}
We do a comparative study by varying the number of clusters (\textit{n-clusters}) obtained with \textit{Spectral clustering} and comparing their prediction performance based on the evaluation schemes discussed before. Specifically, for \textit{n-clusters} as 2, 4, 8, 16, 32, and 64, we present Table~\ref{results-table_cluster_analysis} where we compare their training error in terms of RMSE, cumulative variability of the future trend prediction in terms of its RMS, and the ML model uncertainty in prediction quantified by the RMS of model uncertainty. We also include the correlation of the predicted trend with the past altimeter trend (1993-2022). Additionally, we add another column which provides \textit{Spectral clustering}'s performance scores when evaluated solely with the climate models. This last column reports an average correlation score as derived in 4.b.\ref{leave_one_out}. 
\begin{table}[ht!]
	\centering
	\renewcommand{\arraystretch}{1.25}
	\begin{tabularx}{\textwidth}{|X|X|X|X|X|X|}
		\cline{1-6}
		\multicolumn{1}{|c|}{\textbf{\textit{\small n-clusters}}} & \textbf{\small Training RMSE} & \textbf{\small RMS of future predicted trend (cumulative variability} & 
		\textbf{\small RMS of ML model uncertainty} & \textbf{\small Correlation of predicted trend with past altimeter trend} & 
		\textbf{\small Avg correlation on evaluation with climate models} \\ \cline{1-6}
		2   & 0.62 & 0.99 & 0.23 & 0.45 & 0.68 \\ \hline
		4   & 0.51 & 1.05 & 0.19 & 0.59 & 0.65 \\ \hline
		8  & 0.43 & 1.11 & 0.17 & 0.69 & 0.67 \\ \hline
		16  & 0.34 & 1.5  & 0.17 & 0.67 & 0.62 \\ \hline
		32  & 0.38 & 1.29 & 0.24 & 0.69 & 0.62 \\ \hline
		64  & 0.29 & 1.61 & 0.25 & 0.67 & 0.64 \\ \hline
	\end{tabularx}
    \caption{Table comparing ML with Spectral clustering performance across \textit{n-clusters}: 2, 4, 8, 16, 32, 64. It shows the training error (RMSE), RMS of the future predicted trend, RMS of the ML model uncertainty in prediction, correlation of the future predicted trend with the past altimeter trend, and an average correlation score when evaluated only with the climate model datasets as described in 4.b.\ref{leave_one_out}. All the scores are weighted and the RMSE and RMS measures are in mm/year.}
	\label{results-table_cluster_analysis}
\end{table}

With an increase in the number of clusters, there are fewer data points per cluster, so the training data size for each ML model decreases.  Table~\ref{results-table_cluster_analysis} indicates that the training RMSE decreases with increasing \textit{n-clusters}. This is expected as the training process will tend to overfit more with smaller training data per cluster. The RMS that represents the cumulative variability of the future prediction outcome is observed to increase with the increase in \textit{n-clusters} (except for a small drop for \textit{n-clusters} = 32). Notably, the model uncertainty drops and then increases, especially when working with larger number of clusters like \textit{n-clusters} = 32 or 64, as quantified by the RMS in the third column. For such high \textit{n-clusters}, there is a huge decrease in the training data points per cluster and this can lead to more variance in ML's prediction, reducing its confidence. Higher \textit{n-clusters} show predicted trends to be generally more correlated with the past altimeter trend. The last column based on evaluation with climate models doesn't show a significant performance change with \textit{n-clusters}. The score, however, drops slowly with more \textit{n-clusters}. 
For a qualitative comparison, we plot the predicted trend for 2023-2052 as generated by the ML model with 4, 8, and 16 spectral clusters in Figure~\ref{spectral_4_8_and_16_comparison}. 


While we observe slightly better predictions with 8 spectral clusters (from Table~\ref{results-table_cluster_analysis}), we work extensively with the 4-cluster Spectral clustering setup in order to have a fair comparison with the Domain-specified partition with 4 partitions in our case. Having a higher number of clusters also makes it harder for the domain experts to interpret its physical implications. Additionally, upon examining the prediction maps closely from Figure~\ref{spectral_4_8_and_16_comparison}, it can be noted that the difference across various clusters can mostly be seen in the predicted strength of the trends (higher for higher \textit{n-clusters} which also contributes to its higher RMS), and notably not the general prediction patterns themselves. 

\begin{figure}[htbp]
  \centering
\begin{subfloat}[\large Future predicted trend in mm/year with Spectral clustering - 4 clusters for 2023-2052.]
{\includegraphics[width=0.65\textwidth]{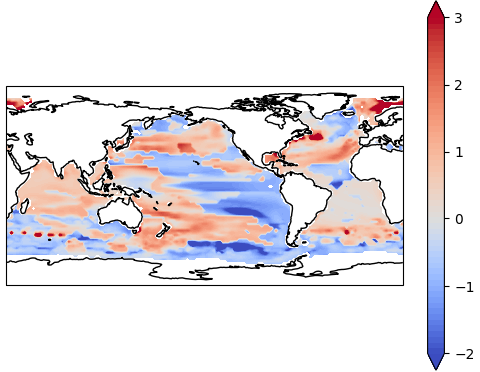}}
  \end{subfloat}
  \caption{}
  \end{figure}
  \clearpage
   \begin{figure}[t!]\ContinuedFloat
    \centering
  \begin{subfloat}[\large Future predicted trend in mm/year with Spectral clustering - 8 clusters for 2023-2052.]
{\includegraphics[width=0.65\textwidth]{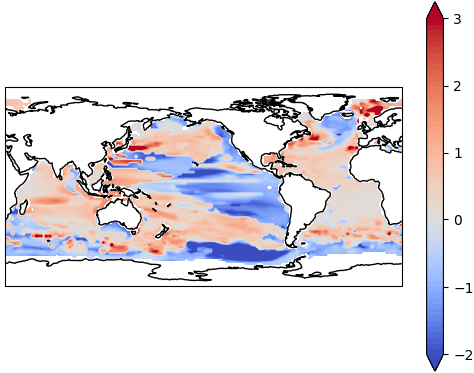}}
  \end{subfloat}
 \begin{subfloat}[\large Future predicted trend in mm/year with Spectral clustering - 16 clusters for 2023-2052.]
{\includegraphics[width=0.65\textwidth]{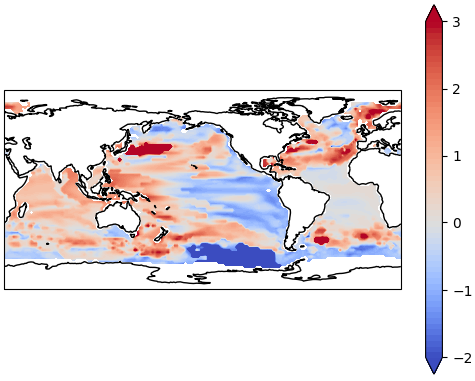}}
  \end{subfloat}
  \caption{Figure shows the ML predicted trend for the future period 2023-2052 with Spectral clustering with (a) 4, (b) 8, and (c) 16 clusters.}
  \label{spectral_4_8_and_16_comparison}
\end{figure}

\section{Discussion}
\label{discussion}
In our framework, fully connected neural networks learn to map climate model projections to altimeter trends. We also present an interpretability study that uses SHAP values to explain the contributions of all the climate models to the final prediction. Spectral clustering shows promise in this application by generating future predictions with ML such that their variability lies in the range we expect, given the variability of the past altimeter observations. These ML predictions have lower uncertainties in impactful areas as shown before. Spectral clustering also shows robustness as it yields better predictions than the one with the Domain-specified partition when evaluated solely with climate models, as described in 4.b.\ref{leave_one_out}. The future predictions are expected to be correlated well with the past altimeter trend and a higher correlation is observed with the predictions obtained from Spectral clustering than the Domain-specified partition.

Our prediction highlights the regional variation in the predicted sea level trend. It is worth noting that the climate model projections used in our framework are "RCP85" and "SSP370" scenarios. These mostly lie in the high level of emission scenarios where a few policies have been put in place to reduce emissions and warming and tackle climate change. Under such circumstances, our prediction outcomes indicate a rising sea level trend around certain important regions such as Japan, India, the South China Sea, the Maritime Continent, Australia, the Gulf Coast and the eastern seaboard of the US, and Mexico. Overall, these predictions suggest that many existing hot spots of sea level rise, including highly populated zones in the western Pacific Ocean and along the US Gulf Coast, will continue to experience rates of sea level rise in excess of the global average. Some of these areas may be limited in their ability to adapt to such changes which could increase the risk of major impacts of sea level rise in the coming decades.   
 
 \section{Conclusion}
We show the effectiveness of neural networks in multi-decadal sea level trend prediction at a 2-degree spatial grid leveraging the projections from climate model large ensembles. We demonstrate that segmenting the spatial grid into partitions employing spectral clustering improves the ML predictions by learning a dedicated ML model per partition. We also supplement our predictions with uncertainty estimates which could be more helpful in interpreting the results. While our framework presents promising results, it is important to note that climate model projections become less certain over time, making long-term predictions based on them challenging. The climate models used in our setup do not incorporate melting ice sheets and their effects on future sea level change~\citep{hamlington2020understanding}. It is pertinent to utilize this to improve the predictions further. The predictions can potentially also improve if we incorporate factors such as wind and temperature, harnessing deep neural networks’ capabilities in handling this diverse data. 
\clearpage
\acknowledgments
This work was supported by NASA Award 80NSSC21K1191. CM was supported in part by a Choose France Chair in AI grant from the French government, administered by INRIA. The author thanks Shivendra Agrawal for their valuable feedback on improving the quality of this work. 
%
%
\datastatement
The climate model output used in this study can be accessed on the Earth System Grid at \url{https://esgf-node.llnl.gov/search/cmip6/}.
The sea level altimeter data from JPL can be accessed at: \url{https://sealevel.nasa.gov/data/dataset/?identifier=SLCP_SEA_SURFACE_HEIGHT_ALT_GRIDS_L4_2SATS_5DAY_6THDEG_V_JPL2205_2205}.

\bibliographystyle{ametsocV6}
\bibliography{references}

\begin{thebibliography}{34}
\providecommand{\natexlab}[1]{#1}
\providecommand{\url}[1]{\texttt{#1}}
\renewcommand{\UrlFont}{\rmfamily}
\providecommand{\urlprefix}{URL }
\expandafter\ifx\csname urlstyle\endcsname\relax
  \providecommand{\doi}[1]{https://doi.org/\discretionary{}{}{}#1}\else
  \providecommand{\doi}{https://doi.org/\discretionary{}{}{}\begingroup \urlstyle{rm}\Url}\fi
\providecommand{\eprint}[2][]{\url{#2}}

\bibitem[{Bahari et~al.(2023)Bahari, Ahmed, Chong, Lai, Huang, Koo, Ng,, and El-Shafie}]{bahari2023predicting}
Bahari, N. A. A. B.~S., A.~N. Ahmed, K.~L. Chong, V.~Lai, Y.~F. Huang, C.~H. Koo, J.~L. Ng, and A.~El-Shafie, 2023: Predicting sea level rise using artificial intelligence: A review. \textit{Archives of Computational Methods in Engineering}, 1--18.

\bibitem[{Balogun and Adebisi(2021)Balogun, and Adebisi}]{balogun2021sea}
Balogun, A.-L., and N.~Adebisi, 2021: Sea level prediction using arima, svr and lstm neural network: assessing the impact of ensemble ocean-atmospheric processes on models’ accuracy. \textit{Geomatics, Natural Hazards and Risk}, \textbf{12~(1)}, 653--674.

\bibitem[{Braakmann-Folgmann et~al.(2017)Braakmann-Folgmann, Roscher, Wenzel, Uebbing,, and Kusche}]{braakmann2017sea}
Braakmann-Folgmann, A., R.~Roscher, S.~Wenzel, B.~Uebbing, and J.~Kusche, 2017: Sea level anomaly prediction using recurrent neural networks. \textit{arXiv preprint arXiv:1710.07099}.

\bibitem[{Danabasoglu et~al.(2020)}]{danabasoglu2020community}
Danabasoglu, G., and Coauthors, 2020: The community earth system model version 2 (cesm2). \textit{Journal of Advances in Modeling Earth Systems}, \textbf{12~(2)}, e2019MS001\,916.

\bibitem[{Dunne et~al.(2013)}]{dunne2013gfdl}
Dunne, J.~P., and Coauthors, 2013: Gfdl’s esm2 global coupled climate--carbon earth system models. part ii: carbon system formulation and baseline simulation characteristics. \textit{Journal of Climate}, \textbf{26~(7)}, 2247--2267.

\bibitem[{Fasullo et~al.(2020{\natexlab{a}})Fasullo, Gent,, and Nerem}]{fasullo2020forced}
Fasullo, J.~T., P.~R. Gent, and R.~Nerem, 2020{\natexlab{a}}: Forced patterns of sea level rise in the community earth system model large ensemble from 1920 to 2100. \textit{Journal of Geophysical Research: Oceans}, \textbf{125~(6)}, e2019JC016\,030.

\bibitem[{Fasullo et~al.(2020{\natexlab{b}})Fasullo, Gent,, and Nerem}]{fasullo2020sea}
Fasullo, J.~T., P.~R. Gent, and R.~S. Nerem, 2020{\natexlab{b}}: Sea level rise in the cesm large ensemble: The role of individual climate forcings and consequences for the coming decades. \textit{Journal of Climate}, \textbf{33~(16)}, 6911--6927.

\bibitem[{Fasullo and Nerem(2018)Fasullo, and Nerem}]{fasullo2018altimeter}
Fasullo, J.~T., and R.~S. Nerem, 2018: Altimeter-era emergence of the patterns of forced sea-level rise in climate models and implications for the future. \textit{Proceedings of the National Academy of Sciences}, \textbf{115~(51)}, 12\,944--12\,949.

\bibitem[{Gal and Ghahramani(2016)Gal, and Ghahramani}]{gal2016dropout}
Gal, Y., and Z.~Ghahramani, 2016: Dropout as a bayesian approximation: Representing model uncertainty in deep learning. \textit{international conference on machine learning}, PMLR, 1050--1059.

\bibitem[{Giorgetta et~al.(2013)}]{giorgetta2013climate}
Giorgetta, M.~A., and Coauthors, 2013: Climate and carbon cycle changes from 1850 to 2100 in mpi-esm simulations for the coupled model intercomparison project phase 5. \textit{Journal of Advances in Modeling Earth Systems}, \textbf{5~(3)}, 572--597.

\bibitem[{Hamlington et~al.(2016)Hamlington, Cheon, Thompson, Merrifield, Nerem, Leben,, and Kim}]{hamlington2016ongoing}
Hamlington, B., S.~Cheon, P.~Thompson, M.~Merrifield, R.~Nerem, R.~Leben, and K.-Y. Kim, 2016: An ongoing shift in pacific ocean sea level. \textit{Journal of Geophysical Research: Oceans}, \textbf{121~(7)}, 5084--5097.

\bibitem[{Hamlington et~al.(2020{\natexlab{a}})Hamlington, Piecuch, Reager, Chandanpurkar, Frederikse, Nerem, Fasullo,, and Cheon}]{hamlington2020origin}
Hamlington, B.~D., C.~G. Piecuch, J.~T. Reager, H.~Chandanpurkar, T.~Frederikse, R.~S. Nerem, J.~T. Fasullo, and S.-H. Cheon, 2020{\natexlab{a}}: Origin of interannual variability in global mean sea level. \textit{Proceedings of the National Academy of Sciences}, \textbf{117~(25)}, 13\,983--13\,990.

\bibitem[{Hamlington et~al.(2020{\natexlab{b}})}]{hamlington2020understanding}
Hamlington, B.~D., and Coauthors, 2020{\natexlab{b}}: Understanding of contemporary regional sea-level change and the implications for the future. \textit{Reviews of Geophysics}, \textbf{58~(3)}, e2019RG000\,672.

\bibitem[{Hassan et~al.(2021)Hassan, Haque,, and Ahmed}]{hassan2021comparative}
Hassan, K. M.~A., M.~A. Haque, and S.~Ahmed, 2021: Comparative study of forecasting global mean sea level rising using machine learning. \textit{2021 International Conference on Electronics, Communications and Information Technology (ICECIT)}, IEEE, 1--4.

\bibitem[{Imani et~al.(2017)Imani, Chen, You, Lan, Kuo, Chang,, and Rateb}]{imani2017spatiotemporal}
Imani, M., Y.-C. Chen, R.-J. You, W.-H. Lan, C.-Y. Kuo, J.-C. Chang, and A.~Rateb, 2017: Spatiotemporal prediction of satellite altimetry sea level anomalies in the tropical pacific ocean. \textit{IEEE Geoscience and Remote Sensing Letters}, \textbf{14~(7)}, 1126--1130.

\bibitem[{Kay et~al.(2015)}]{kay2015community}
Kay, J.~E., and Coauthors, 2015: The community earth system model (cesm) large ensemble project: A community resource for studying climate change in the presence of internal climate variability. \textit{Bulletin of the American Meteorological Society}, \textbf{96~(8)}, 1333--1349.

\bibitem[{Lipovetsky and Conklin(2001)Lipovetsky, and Conklin}]{lipovetsky2001analysis}
Lipovetsky, S., and M.~Conklin, 2001: Analysis of regression in game theory approach. \textit{Applied Stochastic Models in Business and Industry}, \textbf{17~(4)}, 319--330.

\bibitem[{Liu et~al.(2020)Liu, Jin, Wang,, and Xu}]{liu2020sea}
Liu, J., B.~Jin, L.~Wang, and L.~Xu, 2020: Sea surface height prediction with deep learning based on attention mechanism. \textit{IEEE Geoscience and Remote Sensing Letters}.

\bibitem[{Lundberg and Lee(2017)Lundberg, and Lee}]{lundberg2017unified}
Lundberg, S.~M., and S.-I. Lee, 2017: A unified approach to interpreting model predictions. \textit{Advances in neural information processing systems}, \textbf{30}.

\bibitem[{Maher et~al.(2019)}]{maher2019max}
Maher, N., and Coauthors, 2019: The max planck institute grand ensemble: enabling the exploration of climate system variability. \textit{Journal of Advances in Modeling Earth Systems}, \textbf{11~(7)}, 2050--2069.

\bibitem[{Monteleoni et~al.(2011)Monteleoni, Schmidt, Saroha,, and Asplund}]{monteleoni2011tracking}
Monteleoni, C., G.~A. Schmidt, S.~Saroha, and E.~Asplund, 2011: Tracking climate models. \textit{Statistical Analysis and Data Mining: The ASA Data Science Journal}, \textbf{4~(4)}, 372--392.

\bibitem[{M{\"u}ller et~al.(2018)}]{muller2018higher}
M{\"u}ller, W.~A., and Coauthors, 2018: A higher-resolution version of the max planck institute earth system model (mpi-esm1. 2-hr). \textit{Journal of Advances in Modeling Earth Systems}, \textbf{10~(7)}, 1383--1413.

\bibitem[{Nerem et~al.(2018)Nerem, Beckley, Fasullo, Hamlington, Masters,, and Mitchum}]{nerem2018climate}
Nerem, R.~S., B.~D. Beckley, J.~T. Fasullo, B.~D. Hamlington, D.~Masters, and G.~T. Mitchum, 2018: Climate-change--driven accelerated sea-level rise detected in the altimeter era. \textit{Proceedings of the national academy of sciences}, \textbf{115~(9)}, 2022--2025.

\bibitem[{Nieves et~al.(2021)Nieves, Radin,, and Camps-Valls}]{nieves2021predicting}
Nieves, V., C.~Radin, and G.~Camps-Valls, 2021: Predicting regional coastal sea level changes with machine learning. \textit{Scientific Reports}, \textbf{11~(1)}, 1--6.

\bibitem[{Ronneberger et~al.(2015)Ronneberger, Fischer,, and Brox}]{ronneberger2015u}
Ronneberger, O., P.~Fischer, and T.~Brox, 2015: U-net: Convolutional networks for biomedical image segmentation. \textit{International Conference on Medical image computing and computer-assisted intervention}, Springer, 234--241.

\bibitem[{Shapley et~al.(1953)}]{shapley1953value}
Shapley, L.~S., and Coauthors, 1953: A value for n-person games.

\bibitem[{Shi et~al.(2015)Shi, Chen, Wang, Yeung, Wong,, and Woo}]{shi2015convolutional}
Shi, X., Z.~Chen, H.~Wang, D.-Y. Yeung, W.-K. Wong, and W.-c. Woo, 2015: Convolutional lstm network: A machine learning approach for precipitation nowcasting. \textit{Advances in neural information processing systems}, \textbf{28}.

\bibitem[{{Sinha} et~al.(2022){Sinha}, {Monteleoni}, {Fasullo},, and {Nerem}}]{2022AGUFMOS36A..05S}
{Sinha}, S., C.~{Monteleoni}, J.~{Fasullo}, and R.~S. {Nerem}, 2022: {Sea-Level Projections via Spatiotemporal Deep Learning from Altimetry and CESM Large Ensembles}. \textit{AGU Fall Meeting Abstracts}, Vol. 2022, OS36A--05.

\bibitem[{{\v{S}}trumbelj and Kononenko(2014){\v{S}}trumbelj, and Kononenko}]{vstrumbelj2014explaining}
{\v{S}}trumbelj, E., and I.~Kononenko, 2014: Explaining prediction models and individual predictions with feature contributions. \textit{Knowledge and information systems}, \textbf{41}, 647--665.

\bibitem[{Sun et~al.(2020)Sun, Wan,, and Liu}]{sun2020estimation}
Sun, Q., J.~Wan, and S.~Liu, 2020: Estimation of sea level variability in the china sea and its vicinity using the sarima and lstm models. \textit{IEEE Journal of Selected Topics in Applied Earth Observations and Remote Sensing}, \textbf{13}, 3317--3326.

\bibitem[{Vestergaard et~al.(2010)Vestergaard, Nielsen,, and Andersen}]{vestergaard2010seventeen}
Vestergaard, J.~S., A.~A. Nielsen, and O.~B. Andersen, 2010: Seventeen years of global ssh anomalies analyzed by a maximum information based extension to eof analysis. \textit{Power}, \textbf{2}, 1--54.

\bibitem[{Wang et~al.(2022)Wang, Wang, Wu, Liu, Qi, Sun,, and Fu}]{AHybridMultivariateDeepLearningNetworkforMultistepAheadSeaLevelAnomalyForecasting}
Wang, G., X.~Wang, X.~Wu, K.~Liu, Y.~Qi, C.~Sun, and H.~Fu, 2022: A hybrid multivariate deep learning network for multistep ahead sea level anomaly forecasting. \textit{Journal of Atmospheric and Oceanic Technology}, \textbf{39~(3)}, 285 -- 301, \doi{https://doi.org/10.1175/JTECH-D-21-0043.1}, \urlprefix\url{https://journals.ametsoc.org/view/journals/atot/39/3/JTECH-D-21-0043.1.xml}.

\bibitem[{Young(1985)}]{Young1985MonotonicSO}
Young, H.~P., 1985: Monotonic solutions of cooperative games. \textit{International Journal of Game Theory}, \textbf{14}, 65--72, \urlprefix\url{https://api.semanticscholar.org/CorpusID:122758426}.

\bibitem[{Zhao et~al.(2019)Zhao, Fan,, and Mu}]{zhao2019sea}
Zhao, J., Y.~Fan, and Y.~Mu, 2019: Sea level prediction in the yellow sea from satellite altimetry with a combined least squares-neural network approach. \textit{Marine geodesy}, \textbf{42~(4)}, 344--366.

\end{thebibliography}

\end{document}